\documentclass[10pt,twocolumn,letterpaper]{article}

\usepackage{iccv}
\usepackage{times}
\usepackage{epsfig}
\usepackage{graphicx}
\usepackage{amsmath}
\usepackage{amssymb}
\usepackage{enumitem}

\usepackage[breaklinks=true,bookmarks=false]{hyperref}

\iccvfinalcopy % *** Uncomment this line for the final submission

\def\iccvPaperID{4} % *** Enter the ICCV Paper ID here

\begin{document}

%%%%%%%%% TITLE
\title{Fine-Grained Product Class Recognition for Assisted Shopping}

\author{Marian George, Dejan Mircic, G{\'a}bor S{\"o}r{\"o}s, Christian Floerkemeier, Friedemann Mattern\\
Department of Computer Science\\
ETH Zurich\\
{\tt\small \{mageorge,gabor.soros,floerkem,mattern\}@inf.ethz.ch, mircicd@student.ethz.ch}
%\and
}

\maketitle
\thispagestyle{empty}

%%%%%%%%% ABSTRACT
\begin{abstract}
%Providing assistive solutions that target a better shopping experience for users has a direct impact on improving the quality of life of the healthy as well as the visually impaired. In this paper, we present a fine-grained product recognition system that promotes the convenience of locating items on a shopping list in shelves images of grocery stores. First, we automatically recognize useful text on product packaging, e.g., product name and brand, which is then used in inferring the corresponding product class of a text item that the user inputs in their list. We detect text on training images of the fine-grained GroceryProducts dataset that is collected in cross-dataset settings, where training images and test images have different conditions. Second, we discover discriminative patches on product packaging to differentiate between the visually similar product classes and increase robustness against continuous changes in product design. Finally, we continuously improve the recognition accuracy of our system through active learning. Results show the robustness of our method to cross-domain challenges, and scalability to an increasing number of products with minimal re-training.
Assistive solutions for a better shopping experience can improve the quality of life of people, in particular also of visually impaired shoppers. We present a system that visually recognizes the fine-grained product classes of items on a shopping list, in shelves images taken with a smartphone in a grocery store. Our system consists of three components: (a) We automatically recognize useful text on product packaging, e.g., product name and brand, and build a mapping of words to product classes based on the large-scale GroceryProducts dataset. When the user populates the shopping list, we automatically infer the product class of each entered word. (b) We perform fine-grained product class recognition when the user is facing a shelf. We discover discriminative patches on product packaging to differentiate between visually similar product classes and to increase the robustness against continuous changes in product design. (c) We continuously improve the  recognition accuracy through active learning. Our experiments show the robustness of the proposed method against cross-domain challenges, and the scalability to an increasing number of products with minimal re-training.
\end{abstract}

%%%%%%%%% BODY TEXT
\section{Introduction}
Shopping, especially grocery shopping, is a frequent activity in people's life. Smart devices like smartphones and smart glasses provide various opportunities for improving shopping experience. Through applying advanced computer vision tools to images taken with the user's smart device, a better understanding of the user's surrounding environment in a store can be achieved. In this work, we focus on the classification of products on shelves around the user in a grocery store. Recognizing the products that the user is facing can be used in recommending related products, reviewing prices, and assisting the user in navigating inside an unfamiliar store. Assisted navigation in stores is essential for improving the autonomy and independence of the visually impaired in performing their shopping activities.

When populating a shopping list, users frequently write the names or the brands of products instead of their respective classes (e.g., Coca-Cola instead of soft drink). Since our goal is to recognize the product classes, we need to map product names/brands to their respective classes in a scalable and efficient manner with no supervision from the user.

To be applicable in real-world settings, a grocery product recognition system needs to differentiate between a large number of fine-grained product classes. Fine-grained recognition in computer vision targets the problem of sub-ordinate categorization, e.g., recognizing specific classes of birds \cite{ucsd_birds}, cars \cite{cars_stanford}, or flowers \cite{oxford_flowers}. Fine-grained grocery product recognition introduces additional challenges to the general object recognition task. For instance, several product classes are visually similar in the overall shape of the product (e.g., water, juice, and oil-vinegar classes have bottle-shaped products). Also, some product classes possess large variability in the shape of their products which is hard to capture in a single model (e.g.,  cheese and bakery).

Several classification methods rely on the availability of large datasets that cover the objects in various shapes, occlusions, and orientations. The performance of such methods improves by increasing the number of training images that are from the same distribution of the test images \cite{dataset_bias}. However, in real-world applications like in the domain of robotics or assistive systems for the visually impaired, gathering such datasets is a time-consuming task that does not scale well. A product recognition system needs to be robust against cross-domain settings where test images are in different conditions from the training images. Testing images come from different stores with different imaging conditions resulting from the variability of prospect users and imaging devices. An ideal system would need to be trained once and used in multiple stores and scenarios.

In this paper, we address the problem of large-scale fine-grained product recognition in cross-domain settings. The designed method should satisfy the following requirements:
\setitemize{nolistsep,leftmargin=*}
\begin{itemize}
\item \textit{Scalability} to a large number of product classes and product instances; require no or minimum re-training when adding new products to the dataset or changing the packaging of some of the existing products.
\item \textit{Robustness} to cross-domain settings; to be applicable in real-world settings with thousands of supermarkets and millions of users with different characteristics and health conditions.
\item \textit{Autonomy}; automatically recognize the product classes corresponding to strings of product names or brands entered by the user with no supervision on the input.
\item \textit{Runtime efficiency}; the designed solution should be efficient to run within seconds.
\end{itemize}

\begin{figure}[t]
\begin{center}
   \includegraphics[width=0.95\linewidth]{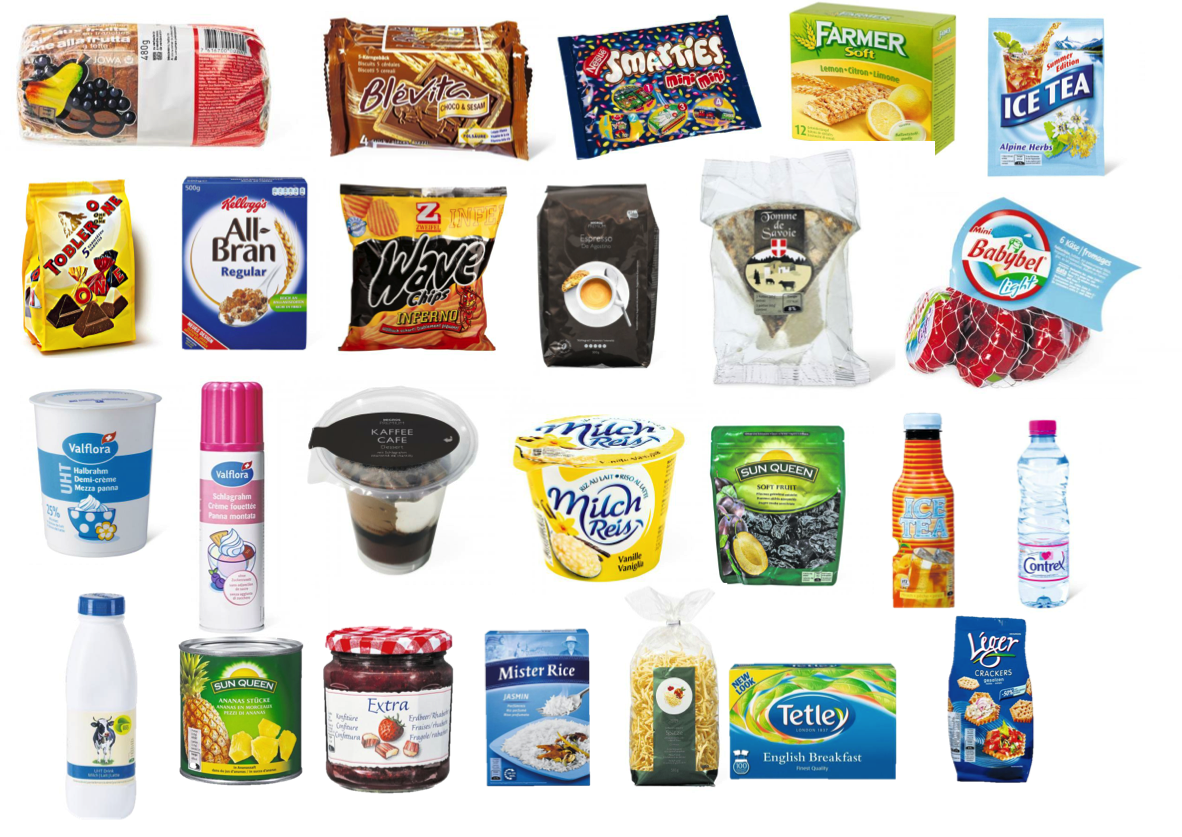}
\end{center}
   \caption{Sample training images from the GroceryProducts \cite{mageorge2014} dataset. Training images are downloaded in ideal studio conditions from the web.}
\label{fig:dataset_training}
\end{figure}

\begin{figure}
\begin{center}
   \includegraphics[width=0.95\linewidth]{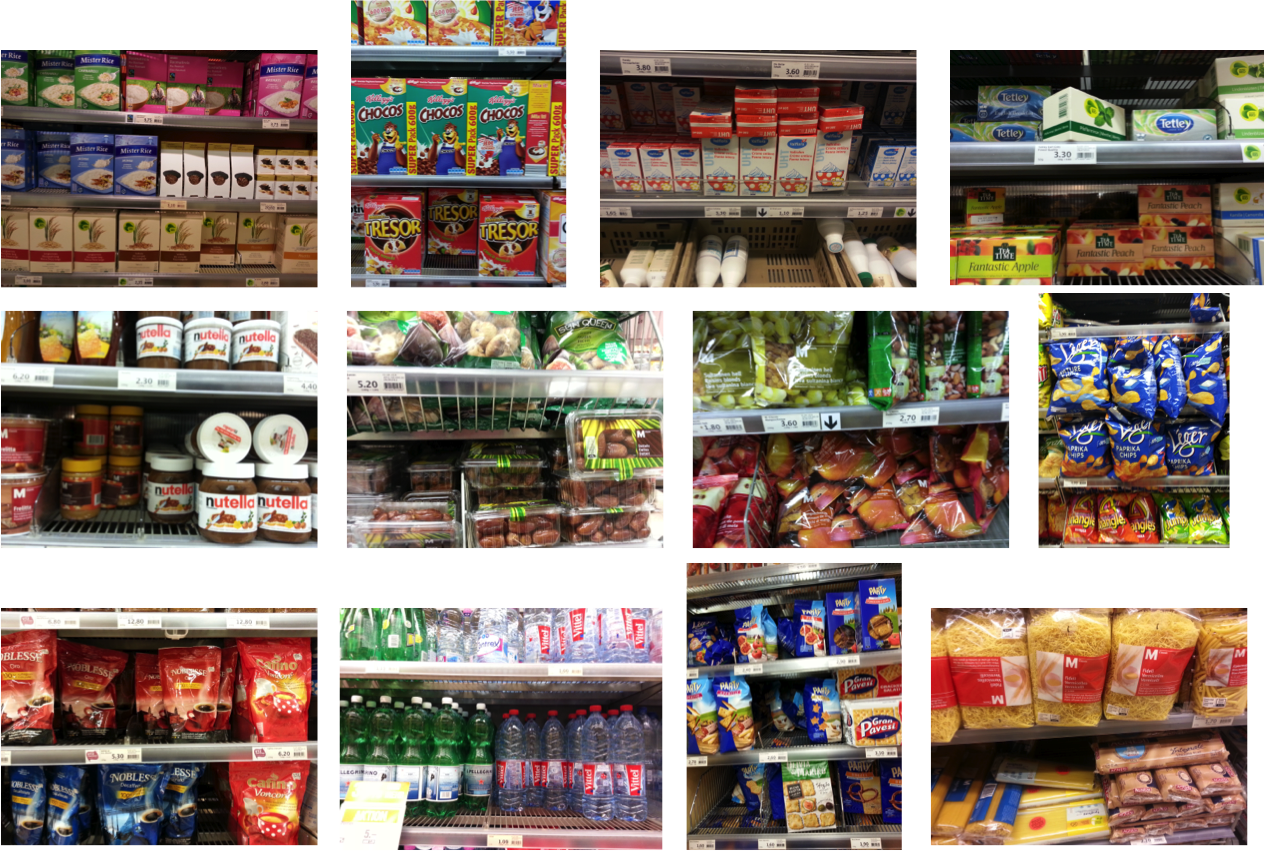}
\end{center}
   \caption{Sample testing images from the GroceryProducts \cite{mageorge2014} dataset. Testing images are taken with a smartphone in real grocery stores with blur, occlusions, and specularities.}
\label{fig:dataset_testing}
\end{figure}

\begin{figure*}[t]
\begin{center}
   \includegraphics[width=\linewidth]{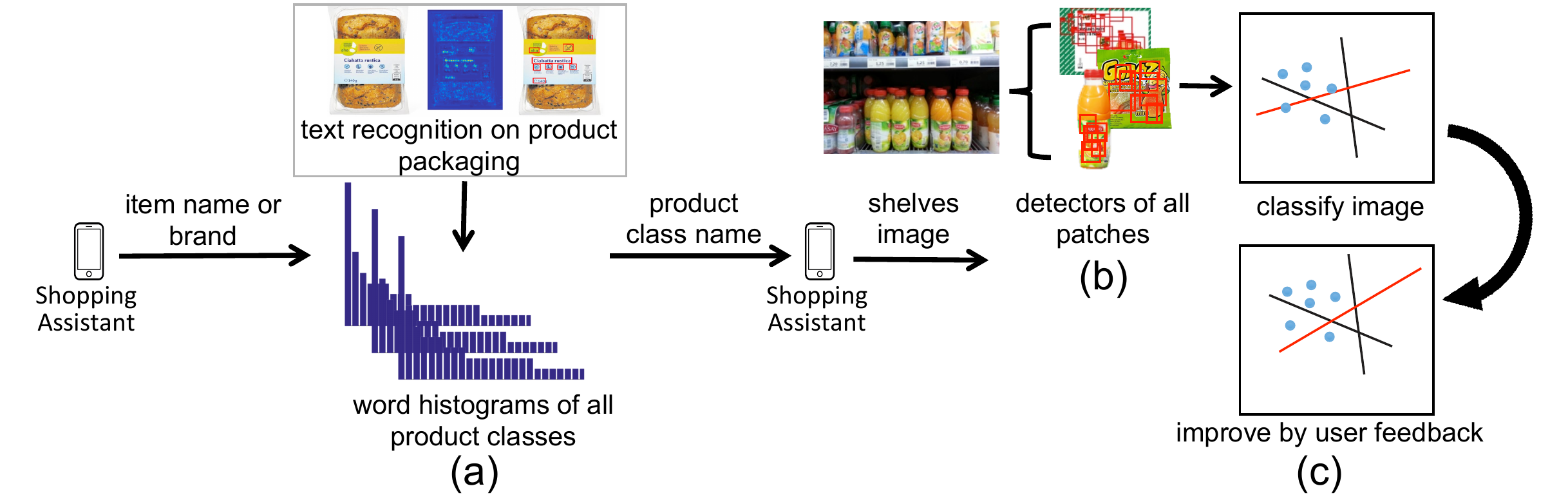}
\end{center}
   \caption{Overview of our system. It consists of three main components: (a) text recognition on product packaging, (b) visual recognition of fine-grained product classes, and (c) recognition improvement by user feedback.}
\label{fig:overview}
\end{figure*}

Our proposed method simultaneously and effectively addresses the above issues. In Figure \ref{fig:overview}, we show an overview of our system, which consists of three components that improve the shopping experience of the user: (a) \textit{Text recognition on product packaging}; automatically recognize useful text on a grocery product packaging like the name and brand of the product using text detection and OCR techniques applied on the training images of grocery products. This information is then used to assist the user by automatically recognizing the product class once a word is entered into the shopping list application. This procedure is scalable to a continuously increasing number of grocery products as it only relies on the ground truth classes of training images without any bounding boxes or additional information from the user. (b) \textit{Product class recognition}; recognize the fine-grained class of a shelves image taken with a smartphone in a real grocery store. Our system works in cross-dataset settings where training images are in different conditions from testing images. We use the GroceryProducts dataset proposed in \cite{mageorge2014}, which contains 26 fine-grained product classes with 3235 training images downloaded in ideal condition from the web, and 680 test images taken with smartphones in real stores. Sample images are shown in Figures \ref{fig:dataset_training} and \ref{fig:dataset_testing} and more details are presented in the following sections. The evaluation of our system shows the effectiveness of discriminative patches in capturing meaningful information on product packaging. (c) \textit{Recognition improvement by user feedback}; continuously improve the accuracy of our system through applying active learning techniques.

%%----------------------------------------------------------------------------------------------------------
\section{Related Work}

Our system is related to the problem of fine-grained object recognition in computer vision. Several approaches have been proposed for recognizing sub-ordinate categories of birds \cite{ucsd_birds, discrim_1, discrim_4, discrim_5}, flowers \cite{oxford_flowers, discrim_3}, and other classes \cite{discrim_2, discrim_6, cars_stanford}. The techniques used in these methods along with the representation of images are significantly different from our target domain of grocery products.

Product recognition has gained increasing interest in the past few years \cite{mageorge2014, products_3, products_4, products_5, products_1,products_2, products_6} due to the fast advancements in computer vision techniques and the availability of computationally powerful mobile devices like smartphones and watches. Image retrieval is used in \cite{products_3} and \cite{products_4} to retrieve images visually similar to a query image. Both the query images and the training images contain a single product from the same dataset. In \cite{products_1}, image retrieval is also targeted but in cross-dataset settings through query object segmentation combined with iterative retrieval. Both the training images and query images contain a single product but with different background conditions. Through segmenting the product, better results are achieved. There are several commercial product search engines for single product recognition, like Google Goggles\footnote{www.google.com/mobile/goggles/} and Amazon Flow\footnote{http://flow.a9.com/} that achieve good performance for planar and textured categories like CD or book covers.

Closely related to our system are approaches that focus on grocery product recognition \cite{mageorge2014, products_5,products_6}. A grocery product dataset of 120 product instances is proposed in \cite{products_5}. Each product is represented by an average of 5.6 training images downloaded from the web and test images are manually segmented from video streams of supermarket shelves. Each test image contains a single segmented product. A baseline approach of SIFT \cite{sift}, color histogram, and boosted Haar-like \cite{viola_jones} feature matching is performed.

In \cite{mageorge2014}, a much larger dataset of 26 grocery product classes  is proposed with 3235 training images of product instances and 680 test images of supermarket shelves. The dataset is gathered in cross-domain settings where training images are downloaded from the web with each image containing a single product instance in ideal studio conditions, while test images contain supermarket shelves taken with a smartphone in real stores. The real-world test images suffer from various image degradations such as blur, occlusions, background objects, and different lighting conditions from the training images. We use this dataset in our experiments. In our earlier paper  \cite{mageorge2014}, we proposed a system that targets instance-level retrieval of specific products present in a test image through a pipelined approach of class ranking, dense pixel matching, and global optimization. We achieved mean average precision of around 23.5\%, which is challenging to apply in assistive tools. Our improved system described in this paper, on the other hand, targets fine-grained product class recognition through automatic discovery of discriminative patches in product images. Additionally, we perform text recognition on product packaging for mapping strings to product classes, and apply active learning to continuously improve the classification performance. The second component of our system is complementary to \cite{mageorge2014}, where we achieve better product classification performance.

A system that targets grocery product detection in video streams is proposed in \cite{products_6}. The system tries to find items on a shopping list in video streams of supermarket shelves. Keypoints in the image are recommended to search for products. The system uses the dataset of 120 products proposed in \cite{products_5}. The authors, however, restrict the search space for each test image to 10 products only, by limiting the number of possible items on a shopping list. Furthermore, they assume that training images of the items on the list are given as an input to the system during query time, which is challenging to scale in real settings with thousands of products. Product detection is performed using na\"{i}ve Bayes classification of SURF \cite{surf} descriptors. Our system is different in several ways. First, we do not restrict the number of items present on the shopping list. Instead, we automatically map each item on the list to its fine-grained product class through our proposed text-recognition-on-product-packaging approach (Section \ref{sec:text_recognition}). Accordingly, our method is scalable to a continuously increasing number of products, where no user input is needed for populating the list. Second, instead of na\"{i}ve Bayes classification, our approach for product recognition relies on discovering discriminative patches on product packaging that differentiate between visually similar products. Finally, we evaluate our system on a much larger dataset, which significantly affects the product recognition accuracy; while each product class in \cite{products_5} is represented by an average of 5.6 images of the \textit{same} specific product, each product class in \cite{mageorge2014} is represented by an average of 112 specific products (each specific product is represented by a \textit{single} image), which is challenging to capture by a single model.

%\textcolor{red}{recognizing products in grocery stores (RFID tags)}  
%%---------------------------------------------------------------------------------------------------------
\section{Product Recognition for Assisted Shopping}
\label{sec:algorithms}

%\textcolor{red}{summarize the shopping list application with features like opportunistically taking images when the phone is in upright position using proximity and orientation sensors, intuitive GUI and non-intrusive notifications. refer to Figure \ref{fig:shopping_list}}

%The whole system can be generalized into two processing pipelines. The first pipeline provides the grocery product classes corresponding to strings of product names or brands entered by the user into the shopping list. Client application sends strings to the server that proceeds to compare them to a precomputed list of keywords produced by our text-recognition-on-product-packaging component (Section \ref{sec:text_recognition}). Figure \ref{fig:shopping_list} shows screenshots from our shopping assistant application.
%The second pipeline classifies images into one of predefined grocery product categories. Images are either opportunistically taken by our application when we detect that the phone is in an upright position through measurements from the proximity and acceleration sensors, or explicitly taken by the user with the smartphone camera. The design and implementation details of our visual-product-class-recognition component are presented in Section \textcolor{red}{\ref{sec:recognition}}.

In Figure \ref{fig:overview}, we give an overview of our system. It consists of \textit{three} components that assist the users in their shopping activities. The first component (Fig. \ref{fig:overview}a) aims at automatically recognizing useful text on grocery product packaging like the name and brand of the product using text detection on the training images of grocery products. We only use the ground truth classes of training images without any bounding boxes or supervision from the user, which is scalable when new products are added. The second component (Fig. \ref{fig:overview}b) recognizes the fine-grained class of an image of supermarket shelves. Training images are in different conditions from testing images, which is more challenging for the recognition process. Images are either opportunistically taken by our application when we detect that the phone is in an upright position through measurements from the proximity and acceleration sensors, or explicitly taken by the user with the smartphone camera. Finally, in the third component (Fig. \ref{fig:overview}c)  we apply active learning to continuously improve the accuracy of our system.

In the following sections, we describe the design and implementation details of each component of our system.
%%----------------------------------------------------------------------------------------------------------
\subsection{Text Recognition on Product Packaging}
\label{sec:text_recognition}
Users usually write the names or brands of products instead of their respective classes (e.g., corn flakes instead of cereal) when populating a shopping list. As our goal is to recognize the product classes, we need to efficiently map words in the list to their respective classes with no supervision from the user.
To achieve this goal, we automatically recognize the text on each product packaging in our training set and compute a histogram to represent how many times each word is encountered in a given class. This histogram is used to measure the confidence of mapping a given word to a corresponding class and is used to rank the possible classes for a given word. 

\begin{figure}
\begin{center}
   \includegraphics[width=\linewidth]{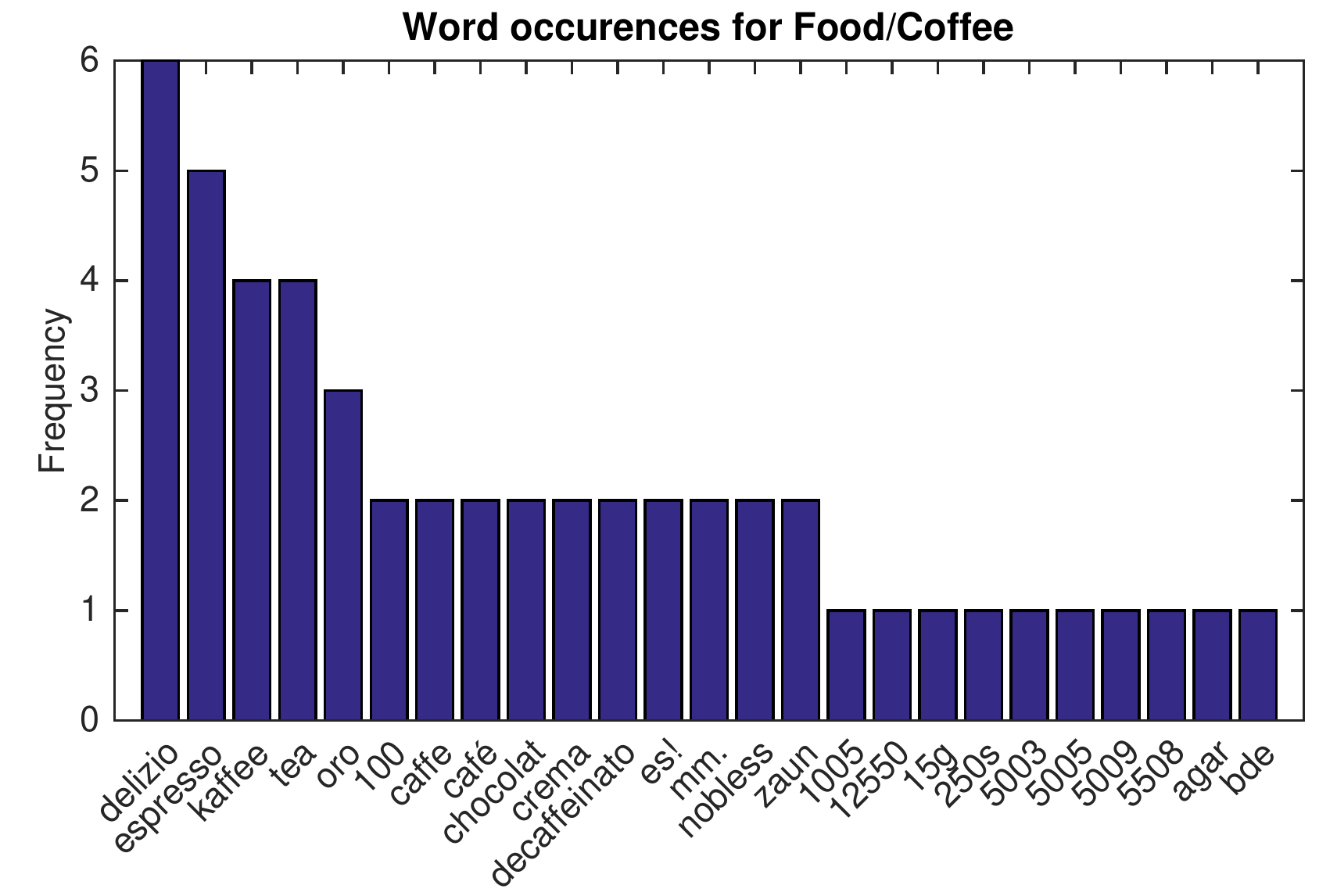}
\end{center}
   \caption{Histogram of word occurrences on the product packaging in the ``Coffee'' category in the dataset.}
\label{fig:word_histogram}
\end{figure}

Recognizing text using Optical Character Recognition (OCR) techniques in natural images requires segmenting text regions from the rest of the image. Applying OCR techniques to whole product images failed to retrieve any useful information. To automatically recognize text regions on each product packaging, we use the approach presented in \cite{jaderberg2014}. The input image first needs to be preprocessed by converting it to grayscale, padding, and normalizing it by subtracting the image mean and dividing by the standard deviation. Then the text/no-text classifier in \cite{jaderberg2014} is applied on the intermediate image. The output of the classifier is a score for each pixel representing how likely it contains text as shown in Figure \ref{fig:overview}a on the top. To create bounding boxes of text regions, we first mask out all pixels with a classification score of 10 or below to leave only high-scored pixels. Following that, we dilate the remaining pixel areas in 6 iterations as the remaining patches are usually of relatively small sizes. Finally, we ignore all regions with a size of 230 pixels or less, since they likely correspond to short or non-meaningful words such as weight declarations, or no text at all. An example of detected bounding boxes on a product are shown in Figure \ref{fig:overview}a on the top.
Once the text regions are segmented, we use the OCR method in \cite{ocr} to recognize the text in each bounding box. A histogram is then built to represent the frequency distribution of words in each class. The histogram of detected words for class ``Coffee'' in our dataset is shown in Figure \ref{fig:word_histogram}.

\begin{figure}
\begin{center}
   \includegraphics[width=0.95\linewidth]{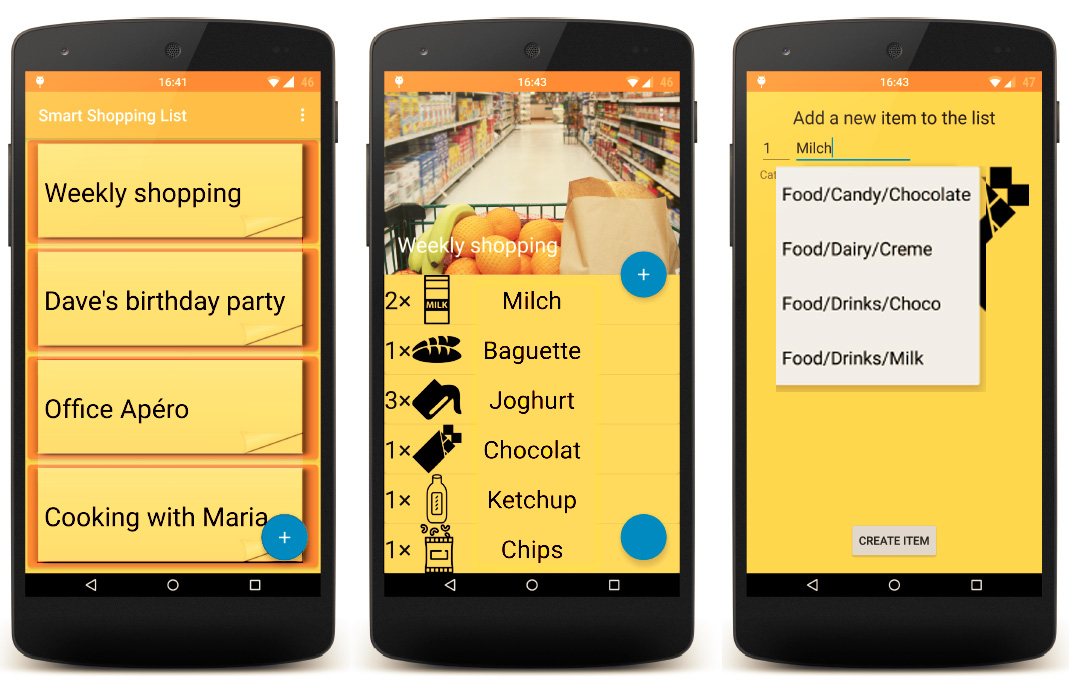}
\end{center}
   \caption{Our shopping assistant. The user enters a textual string that is matched against the pre-computed keyword database, and a filtered list of classes is shown to the user.}
\label{fig:shopping_list}
\end{figure}

When the user writes a product name in the shopping list, the corresponding class is automatically detected if the word occurs in a single class only in the training set. Otherwise, a filtered list of classes ranked by the histogram value is shown to the user to choose from as shown in Figure \ref{fig:shopping_list}. This list typically contains around four classes only, which significantly improves the user experience of populating the shopping list.

%%----------------------------------------------------------------------------------------------------------
\subsection{Product Class Recognition of Shelves Images}
\label{sec:recognition}
Fine-grained grocery product classification poses several challenges as discussed earlier. Such challenges are further aggravated when considering cross-dataset settings in which training images and test images have very different conditions in terms of blur, lighting, deformation, orientation, and the number of products in a given image.

Relying on low-level image features such as SIFT \cite{sift} or HOG \cite{hog} faces difficulty in capturing meaningful image features that are robust against such challenges. The recently proposed mid-level image representations \cite{discriminative_patches,mid_1, mid_2} have achieved impressive results in object and scene classification tasks as they provide a richer encoding of images. Such methods try to discover discriminative patches of a given class, which are patches that occur frequently in the images of the class while they rarely occur in images of other classes. We argue that discriminative patches are beneficial for fine-grained cross-dataset grocery product classification for the following reasons: (1) several product classes may share a common logo. Such image regions are confusing for the classifier and degrades the performance of the system. By extracting discriminative patches from each class, such regions are discarded, which yields better results. (2) While training images are taken in ideal studio conditions, testing images suffer from deformations and occlusions which results in only partial matches between training and testing images. Through relying on features from several image patches instead of whole image, more robust representation is achieved. (3) Several specific product items in a class share common regions, e.g., many rice images contain a rice bowl and many coffee images contain a cup of coffee on the packaging. Capturing such regions and ignoring other less-discriminative regions improves the informativeness of the class model.

\begin{figure}
\begin{center}
   \includegraphics[width=\linewidth]{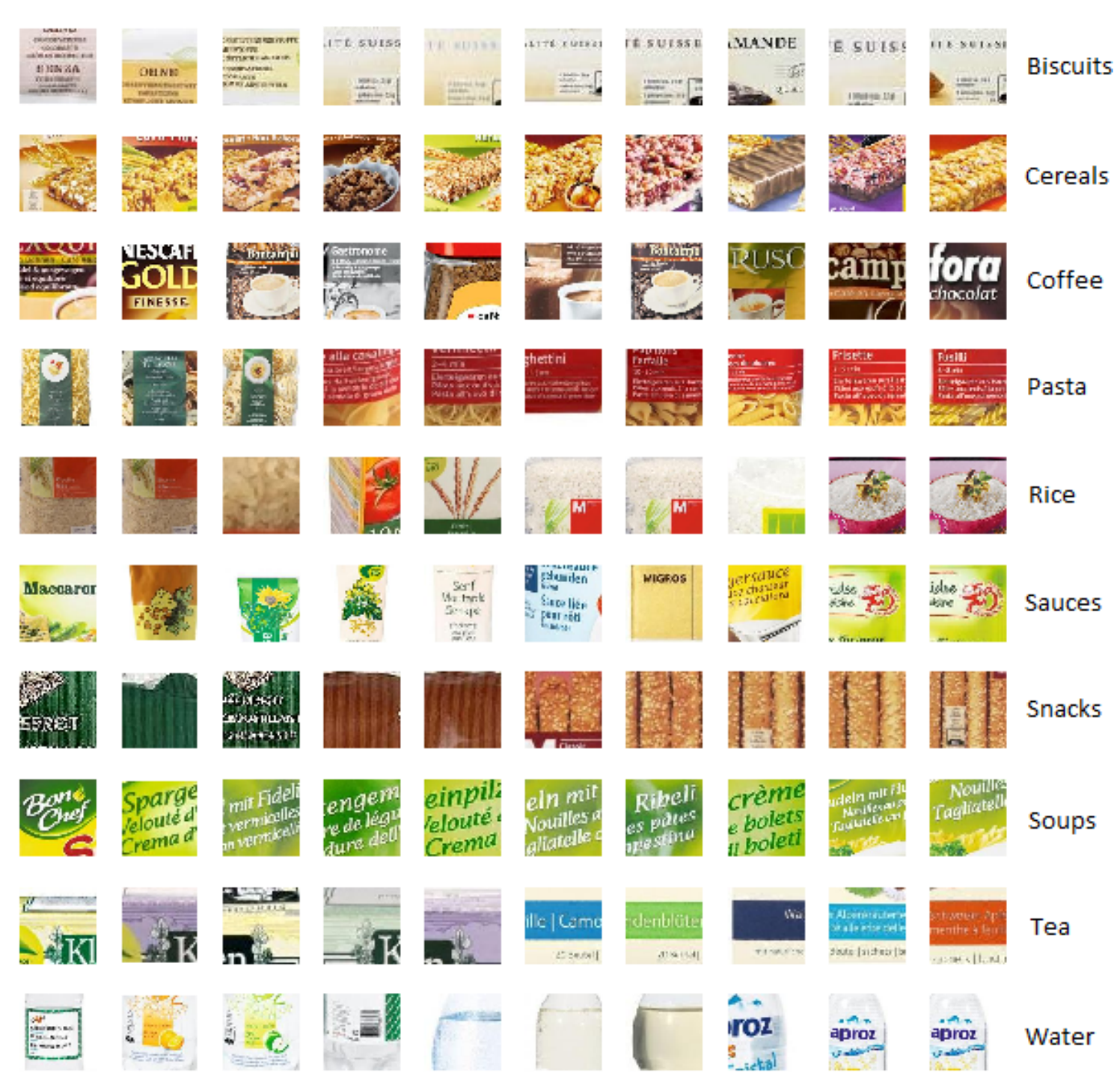}
\end{center}
   \caption{Top 10 discovered discriminative patches for the top 10 correctly classified product classes in the GroceryProducts dataset.}
\label{fig:patches}
\end{figure}

To discover discriminative patches on grocery product packaging, we use the method of \cite{discriminative_patches} to extract mid-level discriminative patches from training images of each grocery product class. The method iterates between clustering and training discriminative SVM detectors. An SVM detector trained on a cluster tries to find similar patches to those in the cluster, which ensures the discriminative property of the cluster. At each step, cross-validation is applied to avoid overfitting. We use the same parameter settings as in \cite{discriminative_patches}. HOG \cite{hog} descriptors of size 8x8 cells with a stride of 8 pixels per cell are computed at 7 different scales. For each class, negative training images are random images from all the other classes in the dataset. The algorithm outputs a few thousand discriminative patches, which are then ranked by the purity and discriminativeness of their clusters. We then take the top 210 patch detectors of each grocery product class to represent each class, as recommended by \cite{discriminative_patches}. Figure \ref{fig:patches} shows the top 10 discriminative patches for the top 10 correctly classified classes in the GroceryProducts dataset. 

The next step is to represent each image by a single feature vector that is suitable for learning a standard SVM classification model. First, we run each patch detector on the whole image. Then, we form a histogram with number of bins equal to the number of classes in the dataset (26 in our case). Each histogram bin contains the highest detection score of the most confident patch of that class, i.e., we do two consecutive steps of max-pooling, first we take the highest score of detecting each of the 210 patches of a class then we take the highest score among all patches. Thus, the histogram has much lower dimensionality than the related ObjectBank \cite{object_bank} descriptor, which makes our descriptor more computationally efficient. Furthermore, our descriptor is not affected by increasing the number of patch detectors per class, as only one value per class is stored in the histogram. To further ensure better runtime performance, we run detectors at a single scale. These histograms are then used to train 1-vs-all linear SVM classifiers for each grocery product class.

To encode spatial information of the extracted features, we use the spatial pyramid image representation \cite{spatial_pyramid}, which has shown significant improvements to the bag-of-words model in object as well as scene classification tasks. We use 2-level spatial pyramid representation. For each image region, we compute the histogram of detection scores described above. Then, we concatenate the histograms from all the image regions, resulting in a histogram of length ${ \tt NumberOfClasses \times (1\times1+2\times2)}$ dimensions. The resulting histograms are then used to train 1-vs-all linear SVM classifiers for each grocery product class, resulting in much improved performance over the whole image histograms as they encode richer information about the spatial arrangement of patches in a given image.

In the evaluation section, we show the superior performance of using discriminative patches in fine-grained product classification over other traditional methods like bag-of-visual-words  \cite{bow} and low-level image features.
%%----------------------------------------------------------------------------------------------------------

\subsection{Adaptive Threshold for User Notification}
We designed our system to be robust against misclassification. This happens for example when the query image contains only background, e.g. floor or ceiling, without any products, or contains product classes that are not in the dataset. Our system only notifies the user of a recognized product if its classification score is higher than a specified threshold. Higher thresholds means that we only notify the user of a product if we are highly confident about the classification result, which results in higher precision at the cost of lower recall values. To find a suitable certainty score, we computed a precision-recall curve when gradually increasing the SVM classification score. In Section \ref{sec:prec-recall}, we perform an analysis of the resulting curve. There are several ways to find a suitable value, e.g. by user satisfaction studies. 
%%----------------------------------------------------------------------------------------------------------

\subsection{Recognition Improvement by User Feedback}
Human users interact constantly with our system, continuously delivering images from the testing domain. These new input images can be used for enhancing the recognition accuracy while maintaining minimal supervision from the user. While our system is generally robust to cross-domain settings, further improvements are expected when images from the test domain are involved in the training process. Active learning \cite{active_learning} allows us to select a subset from the user-provided images to be manually labeled. By selecting the images with the least confident classification score (i.e., those nearest to the learnt SVM hyperplane), the SVM classifier can be re-trained with this additional information to better discriminate the training data. Active learning allows us to select only few images to be labeled, which significantly lowers the amount of manual supervision maintaining high user satisfaction and scalability of our system.
%%----------------------------------------------------------------------------------------------------------

\subsection{Implementation Details}
The shopping assistant has been tested on an LG Nexus 5 running Android Lollipop 5.1. The phone features an 8 MP camera. Images are captured at a resolution of 3264 x 2448 pixels. Featured sensors that are used within the applications are the camera, proximity sensor with two states and accelerometer.

We used the following parameters for our algorithms: the 1-vs-all SVM classifiers were trained using a radial basis function (RBF) kernel with C = 2048 and $\lambda$ = 2. The initial threshold for the discriminative patch detectors was fixed at -1.5.
%%----------------------------------------------------------------------------------------------------------

\section{Evaluation}
%%----------------------------------------------------------------------------------------------------------
\subsection{Dataset}
\label{sec:dataset}
For all our experiments, we use the GroceryProducts dataset from \cite{mageorge2014}. It contains 3235 training images of `Food' grocery products that are organized in 26 hierarchical classes as shown in the axes of Figure \ref{fig:confusion_matrix}.
The number of training images in each fine-grained class varies from 25 to 415 images with an average of 112 images per class. Each specific product instance in a class is represented by exactly one training image, frontal view taken in ideal studio conditions. Figure \ref{fig:dataset_training} shows examples of training images from the dataset.
The testing set is taken with a smartphone in real supermarkets. The testing set consists of 680 annotated test images of supermarket shelves of `Food'  products. Each test image contains multiple products ranging from 6 to 30 products per image. Testing images were taken in real stores and thus have noise and various degradations such as blurriness, occlusions, specularities, and background (e.g., shelves, price tags, ceilings, and floors). Figure \ref{fig:dataset_testing} shows examples of testing images from the dataset.

In all our experiments, we do not rely on any bounding boxes or annotations when classifying testing images to ensure the autonomous behaviour of our system. The ground truth labelling of testing images assumes one class per image, i.e., each testing image contains products from the same fine-grained class. We scale test images to a maximum height of 1080 pixels.

%%----------------------------------------------------------------------------------------------------------
\subsection{Evaluation Metrics}
We evaluate the recognition performance of our system using the average classification accuracy.
To compute the average accuracy over the testing set $D$, we define
\begin{equation}
accuracy_D = \frac{1}{L}\sum_{i=1}^{L}{\frac{k_i}{n_i}},
\end{equation}
where $k_i$ is the number of correctly classified images in class $i$, $L$ is the total number of classes and $n_i$ is the total number of images in class $i$.

%%----------------------------------------------------------------------------------------------------------
%\subsection{Results}
%\textcolor{red}{write some introduction to summarise the section}

\subsection{Classification Performance}
To evaluate the performance of the visual recognition component of our system (Section \ref{sec:recognition}), we compute the average accuracy for the following variants of our system: 
\begin{enumerate}
\item Full: discriminative patches + 2x2 pyramid + SVM
\item DP \& SVM: discriminative patches on whole image + SVM
\item DP \& HS:  discriminative patches + take the class of the patch with highest score as the class of the image (i.e., no SVM training)
\item DP \& 2x2 Pyramid \& HS: discriminative patches + 2x2 pyramid + take the class of the patch with highest score in all 5 regions (1x1+2x2)
\item Baseline: 128-dimensional SURF descriptors quantized by bag-of-words (BoW) model with 200 words + SVM
\end{enumerate}

\begin{table}
\begin{center}
\begin{tabular}{|l|c|}
\hline
Method & Accuracy(\%) \\
\hline\hline
Baseline & 12.4 \\
\hline
DP \& SVM & 41.8\\
DP \& HS & 46.6\\
DP \& 2x2 Pyramid \& HS  & 49.9\\
\hline
\textbf{Full} (DP \& 2x2 Pyramid \& SVM) & \textbf{61.9} \\
%\hline
%DP \& SVM \& BB & 39.41\\
%DP \& HS \& BB & 54.71\\
%DP \& 2x2 Pyramid \& SVM \& BB & \textcolor{red}{xx}   \\
 \hline
\end{tabular}
\end{center}
\caption{Average classification accuracy of different variants of our method and the baseline method on the GroceryProducts dataset.}
\label{table:results}
\end{table}

Table \ref{table:results} shows the average accuracy of the different variants of our system and the baseline method. Our full system achieves an average accuracy of 61.9\%, which significantly outperforms the average accuracy of 12.4\% of the baseline method. Using spatial pyramid representation results in a notable improvement in the performance of our system, where it improves the average accuracy by around 20\%. To examine the quality of the discovered discriminative patches, we report results when using the class of the highest scoring patch among all patches of all classes as the class of the image (DP \& HS). We achieved an average accuracy of 46.6\%, which impressively outperforms using an SVM classifier (DP \& SVM) by around 5\%. Accordingly, the discovered patches are of high quality and represent the data well. The classification accuracy when taking the class of the highest scoring patch in all image regions using 2x2 spatial pyramid (DP \& 2x2 Pyramid \& HS) is inferior to using an SVM classifier (Full), as the histogram used for classification encodes richer image information through spatial context.

%To examine the robustness of our method against the presence of multiple products in a testing image as opposed to only a single product in a training image, we computed the average accuracy when considering the ground truth bounding boxes provided with the test set. We run our algorithm on each bounding box and then take the classification result of the bounding box with the highest classification score as the final result. We achieved an average accuracy of 39.4\% which is slightly better than our result of 37.65\% which shows that our method is robust against the varying conditions between the training and testing images.
%(4) DP \& BB: discriminative patches + bounding boxes), (5) HS \& BB (Discriminative Patches + class of highest score + bounding boxes), (6) DP \& BB \& 2x2 Pyramid: ..., and (7) %Baseline (SURF + BoW + SVM): ...

\begin{figure}
\begin{center}
   \includegraphics[width=\linewidth]{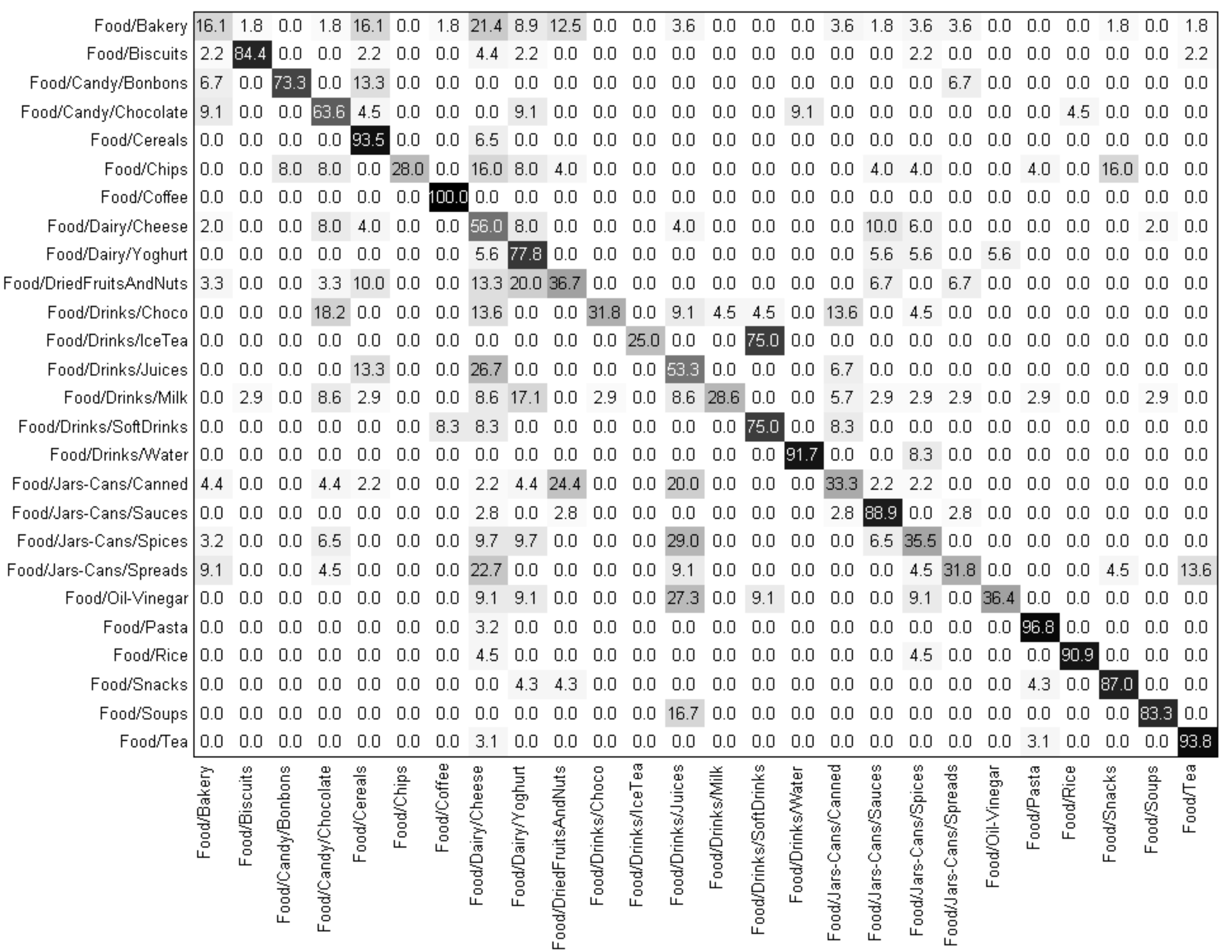}
\end{center}
   \caption{Confusion matrix of the classification results for the 26 fine-grained classes of the GroceryProducts dataset.}
\label{fig:confusion_matrix}
\end{figure}

Figure \ref{fig:confusion_matrix} shows the confusion matrix of classification accuracy over all the 26 classes of the dataset. The top 10 correctly classified classes are \textit{Coffee}, \textit{Pasta}, \textit{Tea}, \textit{Cereals}, \textit{Water}, \textit{Rice}, \textit{Sauces}, \textit{Snacks}, \textit{Biscuits}, and \textit{Soups}. Such classes have distinct product packaging that yield highly discriminative patches as shown in Figure {\ref{fig:patches}}. For example, \textit{Coffee} class is characterised by the cup of coffee on most products, and \textit{pasta} bags usually are transparent showing the uniquely textured pasta inside.

Failure cases include \textit{Bakery}, \textit{Chips}, \textit{Ice Tea}, and \textit{Milk} classes. Reasons for the poor performance varies from one class to the other. For instance, \textit{Bakery} class lacks the presence of logos or discriminative figures on the packaging. Products vary in texture and shape, and are highly deformable which makes it challenging to match training images with testing images. \textit{Chips} packaging is often made up of plastic foil which is prone to reflections and deformations that hinder patch detection. If we inspect the \textit{Ice Tea} class, we observe that it has been misclassified as \textit{Soft Drinks} in 75\% of the test cases. This is explained by the common shape and general appearance of both classes. Also, they often share the same manufacturer which makes it more challenging to differentiate between them. \textit{Milk} class is mostly confused with \textit{Yoghurt} due to similar packaging shape.

\begin{figure}
\begin{center}
   \includegraphics[width= \linewidth]{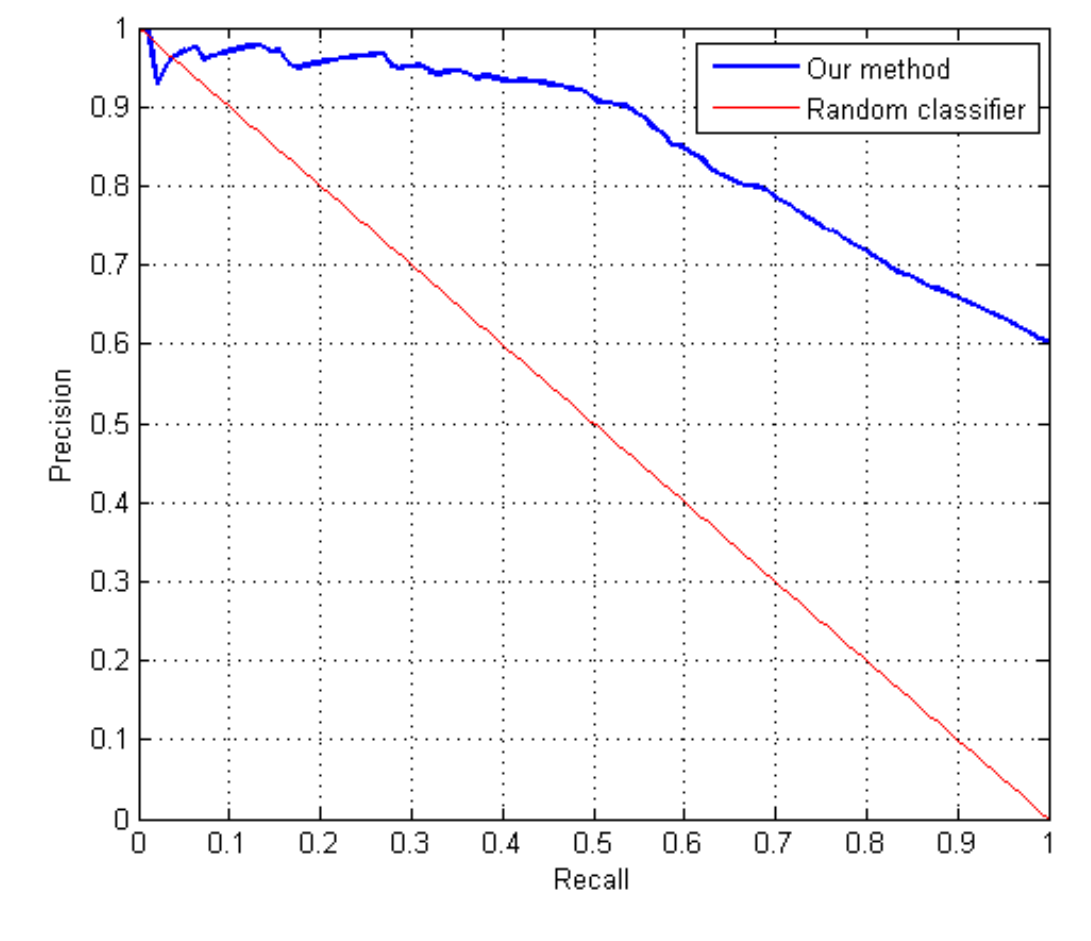}
\end{center} 
   \caption{Precision-recall curve for thresholding the SVM classification score. Our method yields high precision of over 90\% for recall values up to 50\%, as shown by the flatness out our curve.}
\label{fig:prec_recall}
\end{figure}
%%----------------------------------------------------------------------------------------------------------
\subsection{Adaptive Threshold Analysis}
\label{sec:prec-recall}
Figure \ref{fig:prec_recall} shows the precision/recall curve when varying the threshold of the SVM score.  As we tighten the certainty score, the precision increases but at the cost of lower coverage of recognized products (less recall). As can be seen from the graph, our system achieves high precision values of over 90\% for recall values up to 50\% which highlights the usability of our system. Through user satisfaction studies, a suitable threshold can be specified that satisfies user needs and convenience.
%%----------------------------------------------------------------------------------------------------------

\subsection{Active Learning Performance}

\begin{figure}
\begin{center}
   \includegraphics[width=0.98\linewidth]{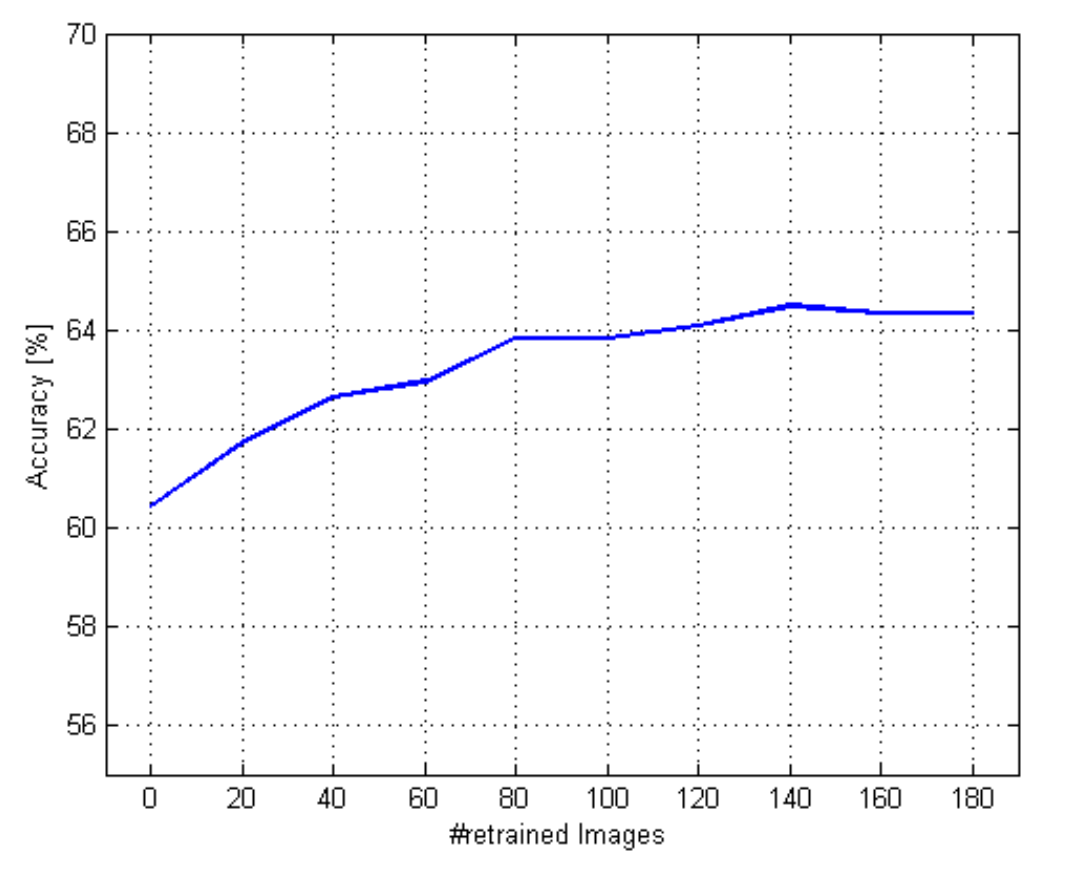}
\end{center}
   \caption{Average classification accuracy for increasing number of images used for learning in the active learning procedure. Testing set size is fixed at 500 images, maximum learning set size is 180 images, and the iteration step size is 20 images.}
\label{fig:active_learning}
\end{figure}

To examine the effectiveness of active learning in improving the recognition accuracy, we ran 3 different experiments where we divide the original testing set of 680 images into 2 disjoint sets: a learning set and a testing set. The testing set is remained fixed and is used to test the performance of the SVM classification. The learning set is used in the iterations of the active learning process, where we gradually increment the number of labeled images from the learning set that are used in re-training the SVM classifier. In each of the 3 experiments, we vary the number of images in the learning set and the testing set. For each iteration in the active learning process, we executed 10 runs with randomly selected learning sets and averaged the accuracy. Figure \ref{fig:active_learning} shows the result for the first experiment with a maximum learning set size of 180 images, a constant testing set size of 500 images, and iteration step size of 20 images. The initial accuracy is 60.5\%. It increases with an increasing learning set up to a size of 140 images, after which it stagnates and stays stable at 64.4\%. Similar behaviors are observed with the other 2 experiments. For the second experiment of a learning set size of up to 280 images, a constant testing set size of 400 images, and iteration step size of 20 images, the accuracy increases again with increased size of the learning set, stabilizes at around 160 images with 65.5\% accuracy, and decreases slightly to 65.2\% after 220 images which can be attributed to the addition of outlier images that confuse the classifier. The  final setting uses a learning set size of up to 500 images, a constant testing set size of 180 images, and iteration step size of 50 images. The initial accuracy with no learned images is relatively low at 56.0\%. It then increases up to 64.0\% with 450 learned images, after which it drops slightly to 63.9\% at 500 learned images.

The experiments show that our active learning procedure succeeds in improving the recognition accuracy due to the addition of more informative images to the training set, with the advantage of minimal supervision to maintain user satisfaction and computational efficiency.
%as follows: (1) 500 images for testing and 180 images for learning, (2) 400 images for testing and 280 images for learning, and (3) 180 images for testing and 500 images for learning.

%%----------------------------------------------------------------------------------------------------------

\subsection{Runtime Performance}
We run our experiments on a machine with Intel Core i7-4770 CPU running at 3.40 GHz and 16 GB RAM without code optimization. Training of each of the 1-vs-all SVMs takes around $16.8$ seconds. Classifying a single image with our proposed method including feature extraction time takes on average $27.6$ seconds and $106.9$ seconds with the additional use of 2x2 spatial pyramids, when using a single thread. The main time consuming task in the classification process is running the patch detectors. Accordingly, the runtime of the classification process can be easily improved through parallelization as the discriminative patch detectors are completely independent.

%%----------------------------------------------------------------------------------------------------------
\section{Conclusion and Future Work}
We presented a product recognition system that recognizes the fine-grained product classes of supermarket shelves images taken with a smartphone in real grocery stores. We have shown that we can improve the user's shopping experience through visual recognition of product classes of items on a shopping list, in addition to unsupervised mapping of textual strings on the shopping list to their corresponding product classes. Through discovering discriminative patches on training product images, we achieve robustness against cross-domain challenges and against continuous design changes in product packaging. In future work, we plan to implement the system on wearable computers like Google Glass to achieve more natural interaction with our system and experiment with different imaging conditions. We will also examine how much improvement in the recognition accuracy can be achieved using a sequence of images from video streams in grocery stores.To be more useful for the visually impaired, we intend to investigate into instance-level retrieval, to recognize the specific product desired by the user.  To improve runtime efficiency, we will work on improving the efficiency of the patch detection algorithm.

{\small
\bibliographystyle{ieee}
\bibliography{egbib}
}

\end{document}